# Microscope 2.0: An Augmented Reality Microscope with Real-time Artificial Intelligence Integration

Po-Hsuan (Cameron) Chen*, Krishna Gadepalli*, Robert MacDonald*, Yun Liu, Kunal Nagpal, Timo Kohlberger, Jeffrey Dean, Greg S. Corrado, Jason D. Hipp, Martin C. Stumpe

Google AI Healthcare, Mountain View, CA, USA.

*These authors contributed equally to this work.

**The brightfield microscope is instrumental in the visual examination of both biological and physical samples at sub-millimeter scales. One key clinical application has been in cancer histopathology, where the microscopic assessment of the tissue samples is used for the diagnosis and staging of cancer and thus guides clinical therapy[1]. However, the interpretation of these samples is inherently subjective, resulting in significant diagnostic variability[2,3]. Moreover, in many regions of the world, access to pathologists is severely limited due to lack of trained personnel[4]. In this regard, Artificial Intelligence (AI) based tools promise to improve the access and quality of healthcare[5–7]. However, despite significant advances in AI research, integration of these tools into real-world cancer diagnosis workflows remains challenging because of the costs of image digitization and difficulties in deploying AI solutions[8,9]. Here we propose a cost-effective solution to the integration of AI: the Augmented Reality Microscope (ARM). The ARM overlays AI-based information onto the current view of the sample through the optical pathway in real-time, enabling seamless integration of AI into the regular microscopy workflow. We demonstrate the utility of ARM in the detection of lymph node metastases in breast cancer and the identification of prostate cancer with a latency that supports real-time workflows. We anticipate that ARM will remove barriers towards the use of AI in microscopic analysis and thus improve the accuracy and efficiency of cancer diagnosis. This approach is applicable to other microscopy tasks and AI algorithms in the life sciences[10] and beyond[11,12].**

Microscopic examination of samples is the gold standard for the diagnosis of cancer, autoimmune diseases, infectious diseases, and more. In cancer, the microscopic examination of stained tissue sections is critical for diagnosing and staging the patient's tumor, which informs treatment decisions and prognosis. In cancer, microscopy analysis faces three major challenges. As a form of image interpretation, these examinations are inherently subjective, exhibiting considerable inter-observer and intra-observer variability[2,3]. Moreover, clinical guidelines[1] and studies[13] have begun to require quantitative assessments as part of the effort towards better patient risk stratification[1]. For example, breast cancer staging requires counting mitotic cells and quantification of the tumor burden in lymph nodes by measuring the largest tumor focus. However, despite being helpful in treatment planning, quantification is laborious and error-prone. Lastly, access to disease experts can be limited in both developed and developing countries[4], exacerbating the problem.

As a potential solution, recent advances in AI, specifically deep learning[14], have demonstrated automated medical image analysis with performance comparable to human experts[2,5,6,15,16]. Research has also shown the potential to improve diagnostic accuracy, quantitation and efficiency by applying deep learning algorithms to digitized whole-slide pathology images for cancer classification and detection[5–7]. However, the integration of these advances to cancer diagnosis is not straightforward because of two primary challenges: image digitization and the technical skills required to utilize deep learning algorithms. First, most microscopic examinations are performed using analog microscopes, and a digitized workflow requires significant infrastructure investments. Second, because of differences in hardware, firmware, and software, the use of AI algorithms developed by others is challenging to use even for experts. As such, actual utilization of AI in microscopy frequently remains inaccessible.

Here, we propose a cost-effective solution to these barriers to entry of AI in microscopic analysis: an augmented optical light microscope that enables real-time integration of AI. We define "real-time integration" as adding the capability of AI assistance without slowing down specimen review or modifying the standard workflow. We propose to superimpose the predictions of the AI algorithm on the view of the sample that the user sees through the eyepiece. Because augmenting additional information over the original view is termed augmented reality, we term this microscope the *Augmented Reality Microscope (ARM)*. Although we apply this technology to cancer diagnosis in this paper, the ARM is application-agnostic and can be utilized in other microscopy applications.

Aligned with ARM's goal to serve as a viable platform for AI assistance in microscopy applications, the ARM system satisfies three major design requirements: spatial registration of the augmented information, system response time, and robustness of the deep learning algorithms. First, AI predictions such as tumor or cell locations need to be precisely aligned with the specimen in the observer's field of view (FOV) to retain the correct spatial context. Importantly, this alignment must be insensitive to small changes in the user's eye position relative to the eyepiece (parallax-free) to account for user movements. Second, although the latest deep learning algorithms often require billions of mathematical operations[17], these algorithms have to be applied in real-time to avoid unnatural latency in the workflow. This is especially critical in applications such as cancer diagnosis, where the pathologist is constantly and rapidly panning around the image. Finally, many deep learning algorithms for microscope images were developed using other digitization methods, such as whole-slide scanners in histopathology[5–7]. We demonstrate that two deep learning algorithms for cancer detection and diagnosis respectively remain accurate when transferred to the ARM. These three core capabilities enable the seamless integration of AI into a traditional microscopy workflow.

We designed and developed the ARM system with three major components: an augmented microscope (Figure 1); a computer with a software pipeline for acquiring the microscope images, running the deep learning algorithms, and displaying results in the microscope in real-time; and a set of trained deep learning algorithms. In this prototype, from an opto-mechanical perspective, the ARM includes a brightfield microscope (Nikon Eclipse Ni-U) augmented with two custom modules (Figure 1). The first module is a camera that captures high resolution images of the current FOV. Relay optics were selected and positioned to ensure that the sample was in focus at both the microscope eyepiece and the camera. The second module is a microdisplay that superimposes digital information into the original optical path.

Parallax-free performance required alignment of the microdisplay to the virtual sample plane within 1 mm. From a computer hardware and software perspective, the ARM includes a computer with a high-speed image grabber (BitFlow CYT) and an accelerated compute unit (NVidia Titan Xp GPU). The ARM system leverages custom software pipelining to maximize utilization of different hardware components for different tasks and to improve responsiveness (Figure 2a). Including the computer, the overall cost of the ARM system is at least an order of magnitude lower than conventional whole-slide scanners, without incurring the workflow changes and delays associated with digitization. Furthermore, due to the modular design of the system, it can be easily retrofitted to most microscopes.

Lastly, the application of the deep learning algorithm comprises two phases: training and inference (Extended Data Figure 3). The training phase involves training an algorithm using a large dataset, while the inference phase involves processing an image with the trained deep learning algorithm. Because the microscope FOV (5120×5120 pixels) is larger than the typical image size used to train deep learning algorithms (smaller than 1000×1000 pixels), exhaustive sliding-window inference is generally required to process the entire FOV. To accelerate inference, we applied the concept of fully-convolutional networks (FCN)[18] to the deep learning architecture InceptionV3[19], which we call InceptionV3-FCN (see Methods). Relative to the original architecture, this modification eliminates 75% of the computation while remaining artifact-free (Figure 2b, Extended Data Figure 4), and can be applied to other architectures by following a few design principles in addition to the standard FCN conversion. The combination of pipelining and FCN improved the latency of the ARM system from 2126 ms to 296 ms and frame rate from 0.94 frames per second (fps) to 6.84 fps (Figure 2c). In our experience, this enables the real-time updating of the augmented information to support a rapid workflow. Furthermore, improvements in compute accelerators will naturally lead to further reductions in latency and increase in frame rate over time.

To investigate the potential of ARM as a platform, we developed and tested deep learning algorithms for two clinical tasks: the detection of metastatic breast cancer in lymph nodes and the identification of prostate cancer in prostate specimens. These tasks affect breast cancer and prostate cancer staging respectively, and thus inform therapy decisions. Figure 3a shows several sample FOVs through the ARM for these tasks.

Next, we verified that these algorithms were robust against differences in image quality and color balance. Specifically, the algorithms were developed using images from a different modality, whole-slide scanners, and applied to images captured in the microscope. We sampled FOVs from lymph node and prostate specimens, blinded to the output of the deep learning algorithms. In total, we selected 1000 FOVs from 50 lymph node slides and 1360 FOVs from 34 prostate slides, using the 10X and 20X objectives (Extended Data Table 1). These "medium power" objectives were selected because they are commonly used to search for regions of interest that can be examined in greater detail at higher magnifications. For lymph node metastases detection, the algorithm achieved an area under the receiver operating characteristic curve (area under ROC, or AUC) of 0.96 (95% confidence interval (CI): 0.93-0.98) at 10X and an AUC of 0.98 (95%CI: 0.96-0.99) at 20X. For prostate cancer detection, the algorithm achieved an AUC of 0.95 (95%CI: 0.94-0.96) at 10X and an AUC of 0.98 (95%CI: 0.97-0.99) at 20X. The ROC curves are shown in Figure 3b.

We have presented a novel augmented microscope with real-time AI capabilities to bridge the gap between AI algorithms and the traditional microscopy workflow. As a proof of concept, we have developed and evaluated deep learning algorithms for two applications: the detection of metastatic breast cancer in lymph nodes and the identification of prostate cancer. Further studies will be required to evaluate the impact of using the ARM in actual clinical workflows and with other microscope models. Because the ARM system is designed to align with the regular microscopy workflow, it will be not be appropriate for all tasks. For example, whole-slide digitization may be more efficient for exhaustive AI-based interpretation of the whole specimen.

However, the ARM system can be used for a range of other applications, whether based on AI algorithms and or solely utilizing the augmented reality capabilities. Other clinical applications that can benefit from the ARM include stain quantification[20], size measurements[21], infectious disease detection (e.g. malaria[22] or tuberculosis[23]), and cell or mitosis counting[24] (Figure 4). Beyond the life science, the ARM can potentially be applied to other microscopy applications such as material characterization in metallurgy[12] and defect detection in electronics manufacturing. In metallurgy, the ARM can improve metallurgical analysis by characterizing the material's microstructure[12] and aid the identification of failure-prone patterns. In electronics manufacturing, an integrated circuit that is defective based on tests is decapped to reveal the die, and examined under a brightfield microscope. In this scenario, the ARM can be used to quickly identify the flaw, such as defective interconnections. To conclude, we anticipate that the ARM will enable the seamless integration of AI into the microscopy workflow and improve the efficiency and consistency of the microscopic examination of cancers and beyond.

# Figures

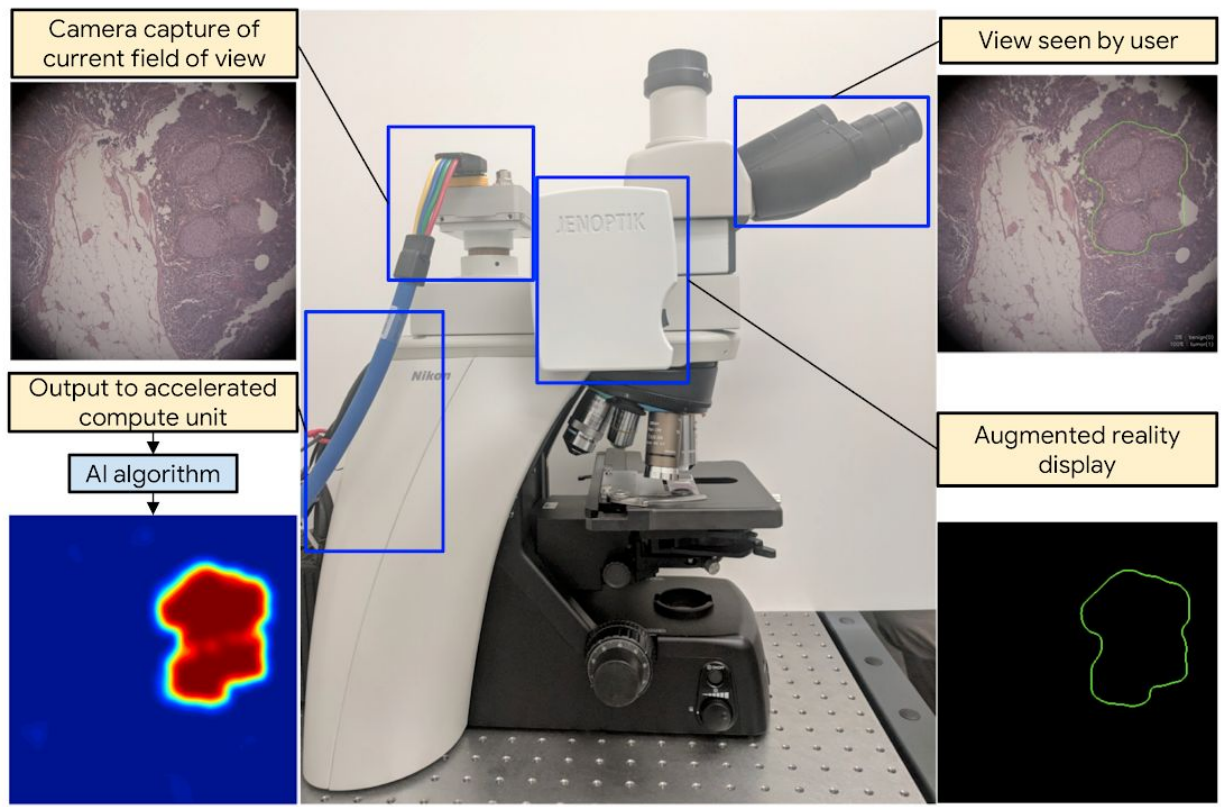

**Figure 1 | Hardware components of the Augmented Reality Microscope (ARM) system enable real-time capture of the field of view and display of information in the eyepiece of the microscope.** The images of the sample are continuously captured. Next, a deep learning algorithm processes each image to produce an inference output (such as a heatmap) with an accelerated compute unit. Finally, the inference output is post-processed to display the most pertinent information without obscuring the original image. For example, outlines of various colors can be used to aid detection and diagnosis tasks, and text such as size measurements can be displayed as well. Technical details can be found in Methods and Extended Data Figure 1.

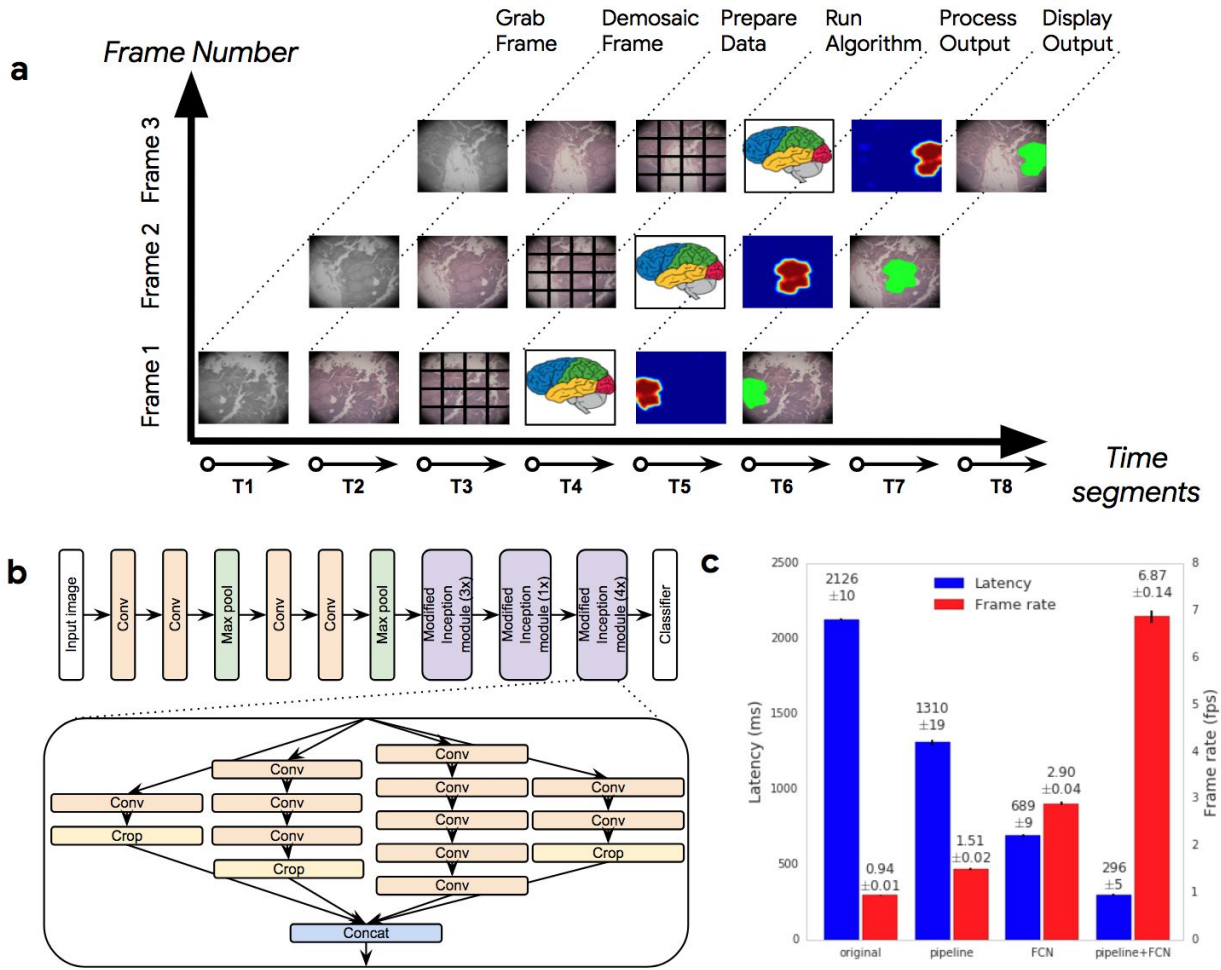

**Figure 2 | Software component of the ARM system improves responsiveness of compute-intensive deep learning algorithms. a,** The process of generating predictions for a single field of view (FOV) requires multiple stages: capturing the FOV, converting that image from the raw sensor data into 3-color RGB pixel values, further image processing, running the deep learning algorithm on the images, processing the output and displaying the processed output. Because each stage has different computing requirements, software pipelining ensures that the appropriate hardware is utilized in each stage and reduces end-to-end latency. **b,** Modifying compute-intensive deep learning architectures to fully-convolutional networks (FCN) reduces compute requirements. The key to doing this without introducing grid-like artifacts is careful cropping (see Methods and Extended Data Figure 4). We chose to modify InceptionV3 as an example because it is the current state-of-the-art[5,6] in breast cancer metastases detection and works well in other medical images: dermatology[16] and ophthalmology[15]. **c,** Latency and frame-rate improvements from pipelining and FCN, and the combination. The error bars represent the standard deviation of 30 measurements of the full sequence of stages for a single FOV. The latency quantified the absolute computational performance of the system, while the frame rate more closely reflects the actual user-perceived responsiveness of the ARM system. For reference, eye tracking studies have shown that for each FOV, pathologists look at several spots, each for 200-400ms[25], indicating that this level of responsiveness is adequate for real-time usage. See also supplementary video for a visual assessment of the responsiveness.

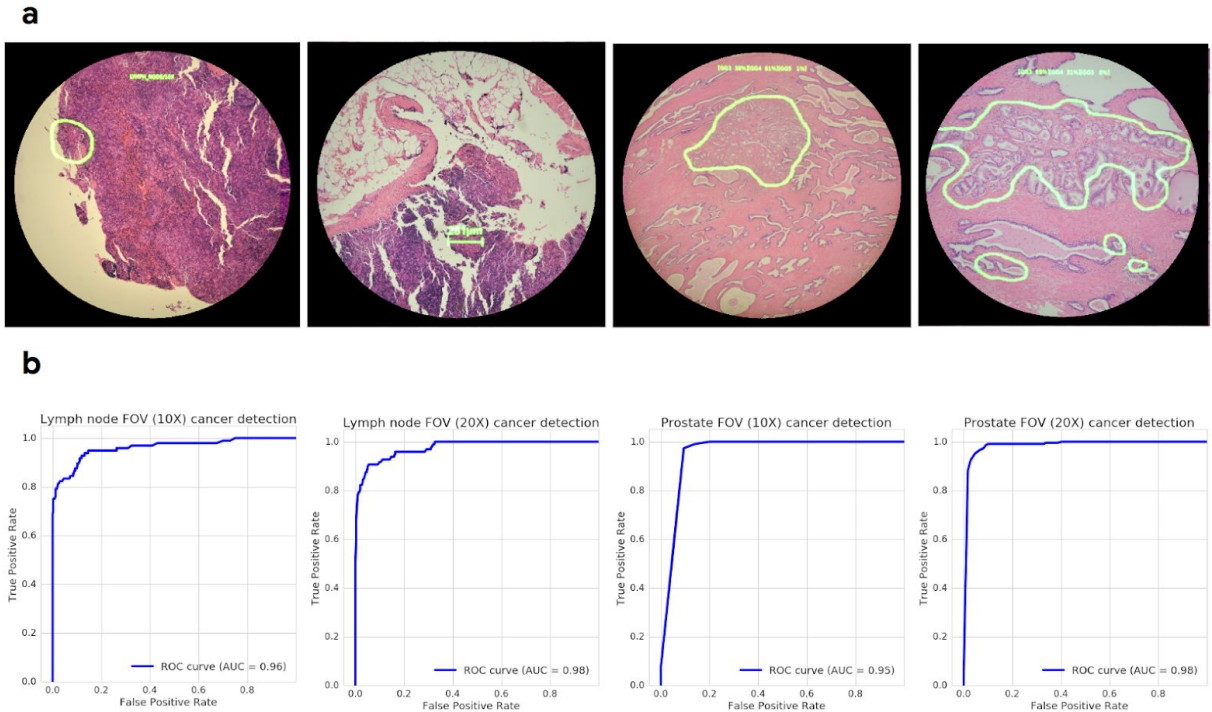

**Figure 3 | Qualitative and quantitative evaluation of lymph node metastases and prostate cancer detection. a,** deep learning algorithm-derived information that can be displayed using the ARM. Left to right: outline of a small metastasis in lymph nodes at medium power to aid detection; numerical measurements of metastasis size in lymph nodes to aid staging; outline of tumor regions in prostatectomies to aid detection; percentage breakdown of Gleason[26] 3,4,5-pattern involvement in the tumor area as a second opinion to aid Gleason grading[27] of prostate samples. Additional examples can be found in Extended Data Figures 5 and 6. **b,** Receiver operating characteristic (ROC) curves to evaluate the accuracy of lymph node metastases detection and prostate cancer detection. For each field of view (FOV), the output from the algorithm is a heatmap depicting the likelihood of cancer at each pixel location. The FOV likelihood is calculated by taking the maximum likelihood across all pixels. The FOV prediction is positive if the FOV likelihood is greater than or equal to a chosen threshold, and negative otherwise. By varying the threshold between 0 and 1, we generate the ROC curve for the true positive rate against the false positive rate. The AUC is defined as the area under the ROC curve. We report the corresponding performance metrics for high accuracy, precision, and recall in Extended Data Table 2.

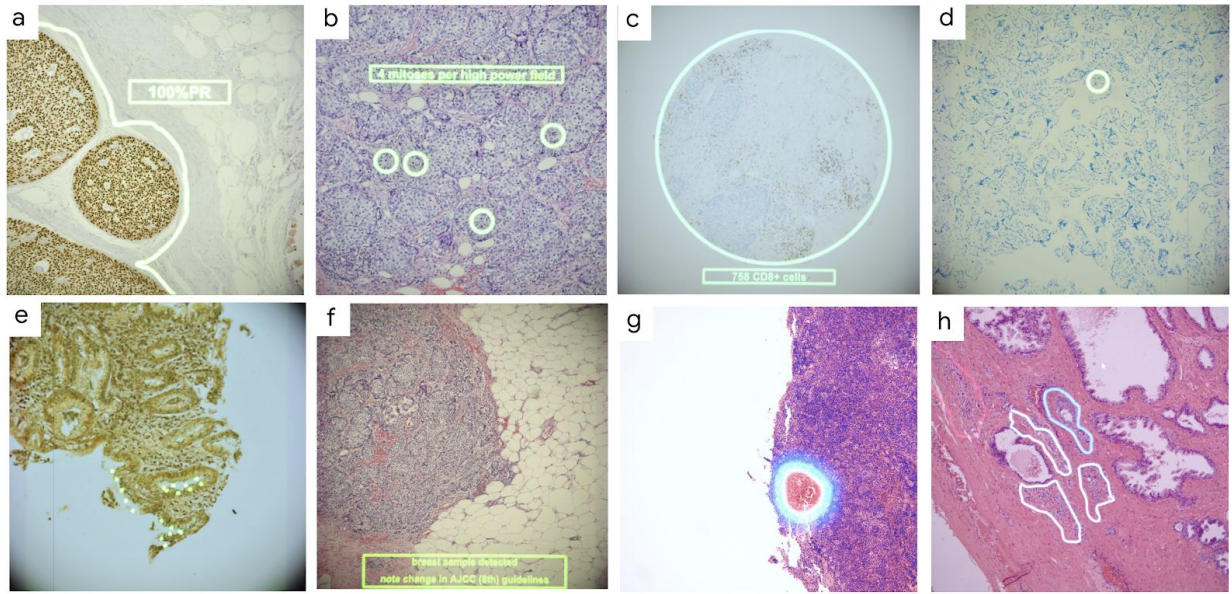

**Figure 4 | Sample future applications that leverage the ARM system's capabilities. a-c,** showcases AI-based stain quantification, mitosis counting, and cell counting respectively. **d-e,** showcases AI-assisted detection for microorganisms (*Mycobacterium tuberculosis* in a sputum smear and *Helicobacter pylori* in a tissue section respectively). **f,** showcases the ability to display notifications. **g,** showcases the ability to display multi-colored images, such as colored contours to convey different levels of uncertainty. **h,** showcases the ability to outline predictions of multiple categories simultaneously, such as Gleason patterns 3, 4, and 5.

# References

1. Amin, M. B. *et al.* The Eighth Edition AJCC Cancer Staging Manual: Continuing to build a bridge from a population-based to a more 'personalized' approach to cancer staging. *CA: A Cancer Journal for Clinicians* **67,** 93–99 (2017).

2. Elmore, J. G. *et al.* Diagnostic concordance among pathologists interpreting breast biopsy specimens. *JAMA* **313,** 1122–1132 (2015).

3. Brimo, F., Schultz, L. & Epstein, J. I. The value of mandatory second opinion pathology review of prostate needle biopsy interpretation before radical prostatectomy. *J. Urol.* **184,** 126–130 (2010).

4. Wilson, M. L. *et al.* Access to pathology and laboratory medicine services: a crucial gap. *Lancet* (2018). doi:10.1016/S0140-6736(18)30458-6

5. Liu, Y. *et al.* Detecting Cancer Metastases on Gigapixel Pathology Images. *Preprint at https://arxiv.org/abs/1703.02442* (2017).

6. Ehteshami Bejnordi, B. *et al.* Diagnostic Assessment of Deep Learning Algorithms for Detection of Lymph Node Metastases in Women With Breast Cancer. *JAMA* **318,** 2199–2210 (2017).

7. Saltz, J. *et al.* Spatial Organization and Molecular Correlation of Tumor-Infiltrating Lymphocytes Using Deep Learning on Pathology Images. *Cell Rep.* **23,** 181–193.e7 (2018).

8. Evans, A. J. *et al.* US Food and Drug Administration Approval of Whole Slide Imaging for Primary Diagnosis: A Key Milestone Is Reached and New Questions Are Raised. *Arch. Pathol. Lab. Med.* (2018). doi:10.5858/arpa.2017-0496-CP

9. Madabhushi, A. & Lee, G. Image analysis and machine learning in digital pathology: Challenges and opportunities. *Med. Image Anal.* **33,** 170–175 (2016).

10. Christiansen, E. M. *et al.* In Silico Labeling: Predicting Fluorescent Labels in Unlabeled Images. *Cell* **173,** 792–803.e19 (2018).

11. Yang, S. J. *et al.* Assessing microscope image focus quality with deep learning. *BMC Bioinformatics* **19,** 77 (2018).

12. Bulgarevich, D. S., Tsukamoto, S., Kasuya, T., Demura, M. & Watanabe, M. Pattern recognition with machine learning on optical microscopy images of typical metallurgical microstructures. *Sci. Rep.* **8,** 2078 (2018).

13. Salgado, R. *et al.* The evaluation of tumor-infiltrating lymphocytes (TILs) in breast cancer: recommendations by an International TILs Working Group 2014. *Ann. Oncol.* **26,** 259–271 (2015).

14. LeCun, Y., Bengio, Y. & Hinton, G. Deep learning. *Nature* **521,** 436–444 (2015).

15. Gulshan, V. *et al.* Development and Validation of a Deep Learning Algorithm for Detection of Diabetic Retinopathy in Retinal Fundus Photographs. *JAMA* **316,** 2402–2410 (2016).

16. Esteva, A. *et al.* Dermatologist-level classification of skin cancer with deep neural networks. *Nature* **542,** 115–118 (2017).

17. He, K., Zhang, X., Ren, S. & Sun, J. Deep Residual Learning for Image Recognition. in *2016 IEEE Conference on Computer Vision and Pattern Recognition (CVPR)* 770–778 (2016).

18. Shelhamer, E., Long, J. & Darrell, T. Fully Convolutional Networks for Semantic Segmentation. *IEEE Trans. Pattern Anal. Mach. Intell.* **39,** 640–651 (2017).

19. Szegedy, C., Vanhoucke, V., Ioffe, S., Shlens, J. & Wojna, Z. Rethinking the inception architecture for computer vision. in *Proceedings of the IEEE Conference on Computer Vision and Pattern Recognition* 2818–2826 (2016).

20. Vandenberghe, M. E. *et al.* Relevance of deep learning to facilitate the diagnosis of HER2 status in breast cancer. *Sci. Rep.* **7,** 45938 (2017).

21. Russ, J. C. *Computer-Assisted Microscopy: The Measurement and Analysis of Images*. (Springer Science & Business Media, 2012).

22. Pirnstill, C. W. & Coté, G. L. Malaria Diagnosis Using a Mobile Phone Polarized Microscope. *Sci.*

*Rep.* **5,** 13368 (2015).

23. Quinn, J. A. *et al.* Deep Convolutional Neural Networks for Microscopy-Based Point of Care Diagnostics. *Proc. Int. Conf. Machine Learning for HealthCare* (2016).

24. Xie, W., Noble, J. A. & Zisserman, A. Microscopy cell counting and detection with fully convolutional regression networks. *Computer Methods in Biomechanics and Biomedical Engineering: Imaging & Visualization* **6,** 283–292 (2018).

25. Krupinski, E. A. *et al.* Eye-movement study and human performance using telepathology virtual slides: implications for medical education and differences with experience. *Hum. Pathol.* **37,** 1543–1556 (2006).

26. Gleason, D. F. & Mellinger, G. T. Prediction of prognosis for prostatic adenocarcinoma by combined histological grading and clinical staging. *J. Urol.* **111,** 58–64 (1974).

27. Epstein, J. I., Allsbrook, W. C., Jr, Amin, M. B., Egevad, L. L. & ISUP Grading Committee. The 2005 International Society of Urological Pathology (ISUP) Consensus Conference on Gleason Grading of Prostatic Carcinoma. *Am. J. Surg. Pathol.* **29,** 1228–1242 (2005).

28. Abadi, M. et al. TensorFlow: A System for Large-Scale Machine Learning. *12th USENIX Symposium on Operating Systems Design and Implementation* (2016).

29. Gutman, D. A. *et al.* Cancer Digital Slide Archive: an informatics resource to support integrated in silico analysis of TCGA pathology data. *J. Am. Med. Inform. Assoc.* **20,** 1091–1098 (2013).

30. van Der Laak, J. A., Pahlplatz, M. M., Hanselaar, A. G. & de Wilde, P. C. Hue-saturation-density (HSD) model for stain recognition in digital images from transmitted light microscopy. *Cytometry* **39,** 275–284 (2000).

## Acknowledgments

We would like to thank the pathologists that provided initial user feedback: Dr. Mahul Amin, Dr. Scott Binder, Dr. Trissia Brown, Dr. Michael Emmert-Buck, Dr. Isabelle Flament, Dr. Niels Olson, Dr. Ankur Sangoi, Dr. Jenny Smith, as well as colleagues who provided assistance with engineering components and paper writing: Toby Boyd, Allen Chai, Lina Dong, William Ito, Jay Kumler, Tsung-Yi Lin, Craig Mermel, Melissa Moran, Robert Nagle, Dave Stephenson, Sujeet Sudhir, Dan Sykora, Matt Weakly.

## Author contributions

P.C. led the deep learning algorithm development and evaluation, K.G. led the software integration, R.M. led the optics development, Y.L. prepared data for the lymph node metastases algorithm, K.N. prepared data for the prostate cancer algorithm, T.K. prepared data for the optical focus assessment algorithm, J.D. and G.C. provided strategic guidance, J.H. provided clinical guidance, M.S. conceived the idea and led the overall development. All authors contributed to writing the manuscript.

# Methods

**Opto-Mechanical Design.** The schematic of the optic design is shown in Extended Data Figure 1. Component design and selection were driven by final performance requirements. The camera and display devices were chosen for effective cell and gland level feature representation. The camera (Adimec S25A80) included a 5120×5120 pixel color sensor with high sensitivity and global shutter capable of capturing images at up to 80 frames/sec. Camera images were captured by an industrial frame-grabber board (Cyton CXP-4) with PCI-E interface to the computer. The microdisplay (eMagin SXGA096, 1292×1036 pixels) was mounted on the side of the microscope and imaged with an achromatic condenser (Nikon MBL71305) at a location tuned to minimize parallax and ensure that the specimen and the display image were simultaneously in focus. The microdisplay includes an HDMI interface for receiving images from the computer. Due to the limited brightness of this display, the second beam splitter (BS2) was chosen to transmit 90% of the light from the display and 10% from the sample, which resulted in good contrast between PI and SI when operating the microscope light source at approximately half of its maximum intensity. The opto-mechanical design used here can be easily retrofitted into most standard bright field microscopes.

**Software and Hardware System.** The application driving the entire system runs on a standard off-the-shelf computer with a BitFlow frame grabber connected to a camera for live image capture and an NVidia Titan Xp GPU for running deep learning algorithms. The process from frame grabbing to the final display is shown in Figure 2. To improve responsiveness, the system is implemented as a highly optimized, pipelined, multi-threaded process, resulting in low overall latency. The software is written in C++ and TensorFlow[28].

The primary pipeline consists of a set of threads that continuously grab an image frame from the camera, debayer it (i.e. convert the raw sensor output into an RGB color image), prepare the data, run the deep learning algorithm, process the results, and finally display the output. Other preprocessing steps such as flat-field correction and white balancing can be done in this thread as well for cameras which cannot do them directly on-chip. To reduce the overall latency, these steps run in parallel for a sequence of successive frames, i.e. the display of frame 'N', generation of heatmap of frame 'N+1', and running algorithm on frame 'N+2' all happen in parallel (Figure 2).

In addition to this primary pipeline, the system also runs a background control thread. One purpose of this thread is to determine whether the camera image is sufficiently in focus to yield accurate deep learning algorithm results. The system uses a convolutional neural network based out-of-focus detection algorithm to assess focus quality. A second purpose of this thread is to determine the currently used microscope objective, so that the deep learning algorithms tuned for the respective magnification is used. An automated detection of the current microscope objective was implemented, without the need to manually specify the magnification. Additionally, settings for white balance and exposure time on the camera can be set to optimal profiles for the respective lens.

**Out-of-focus Detection.** We developed an out-of-focus (OOF) detection algorithm using the proposed InceptionV3-FCN. The network was trained on 216,000 image patches randomly chosen from tissue regions of 27,000 whole slide images digitized with an Aperio AT2 (pixel size 0.252×0.252 μm). Each patch was labelled by three independent non-pathologist human raters to be either in-focus or OOF.

**Datasets for Deep Learning Algorithm Development.** For breast cancer metastases detection, we obtained training data from the Cancer Metastases in Lymph Nodes (Camelyon) 2016 challenge data set[6]. This data set comprises 215 whole slide images from slides digitized by one of two whole-slide scanners: a 3DHISTECH Pannoramic 250 Flash II (pixel size 0.243×0.243μm) or a Hamamatsu XR C12000 (pixel size 0.226×0.226μm). The dataset contains pixel-level ground truth diagnoses of tumor and benign. For prostate cancer identification, we obtained 75 radical prostatectomy slides from The Cancer Genome Atlas (TCGA)[29] and another 376 radical prostatectomy slides from another source. The slides were digitized with an Aperio AT2 scanner (pixel size 0.252×0.252μm). These whole slide images were annotated by pathologists by outlining regions as benign, Gleason pattern 3, Gleason pattern 4 or Gleason pattern 5.

**Datasets for Deep Learning Algorithm Evaluation.** To evaluate the deep learning algorithm performance, we obtained fields of view (FOVs) from 50 slides for lymph node and 34 slides for prostate from two independent sources as testing data (Extended Data Table 1). Each slide came from a different patient case. In total, we collected 1000 FOVs for lymph node slides and 1360 FOVs for prostate slides.

For the lymph node slides, we selected FOVs to represent several categories of benign tissue: capsule and subcapsular sinus, medullary sinuses, adipose tissue, lymphocytes, follicles, broken edges of tissue, and regions with artifacts. The reference standard labels for these slides were established by reviews from two pathologists and adjudication by a third, using pan-cytokeratin (AE1/AE3) immunohistochemistry staining for reference. We collected FOVs from 10 locations, including a maximum of 5 tumor-containing locations where available. For each location, we collected 2 FOVs, using the 20X and 10X objectives respectively.

For the prostate slides, we selected FOVs to represent a diversity of histopathological processes ranging from benign, to inflammation (including prostatitis), to premalignant to various tumor histological Gleason patterns (3, 4, and 5). The reference standard labels for these slides were established by reviews from three pathologists, using PIN4 immunohistochemistry staining where available. Similar to the lymph node dataset, we collected 20 locations (40 FOVs) per slide, including a maximum of 10 tumor-containing locations (20 FOVs) where available.

**Color Variability in the Datasets.** The color distributions of stained tissue slides can vary widely because of variability in factors such as tissue processing, staining protocol, and image capturing device. To quantify the degree of variability in the test set, we plotted quantitative measures of the color (hue, saturation, and a measure of brightness) for all training slides from whole-slide scanners and test images from the ARM microscope. The color distribution exhibited marginal overlap in the lymph node training and test datasets, and minimal overlap in the prostate training and test datasets (Extended Data Figure 7 and Extended Data Figure 8).

**Deep Learning Algorithm Design.**

*Constraints of large image size:* The deep learning component of the ARM system aims to convey the algorithm interpretation of the current field of view (FOV), which is 5120x5120 pixels in this prototype. This size depends on the camera's sensor, and high resolutions yield crisper images. However, this large image size presents a computational challenge for the latest deep learning algorithms, which contain millions of parameters and require billions of floating point operations even for images of size 300 pixels. Because the number of mathematical operations typically scale (approximately) proportionately with the number of input pixels, increasing the input image width and height by a factor of 10 increases the compute requirements by a factor of 100. Thus, developing deep learning algorithms to directly deal with images of such size is currently intractable. Together with our goal of presenting the ARM system as a platform that can utilize other deep learning algorithms, we decided on an alternative, the patch-based approach.

*Limitations of standard patch-based approach:* The patch-based approach crops the input image into smaller patches for training and applies the algorithm in a sliding window across these patches for inference. In histopathology, the interpretation of a region of interest (ROI) generally involves looking at areas of the image near the ROI for additional context; this holds true for both human experts and deep learning algorithms. Because of this requirement, application of the deep learning algorithm for a center ROI of size $x$ generally involves feeding as input a patch of size greater than $x$; and patch-based inference involves re-processing the overlapping "context" regions for nearby patches (Extended Data Figure 2). Thus, the standard patch-based approach, although feasible at training (smaller patch size), results in poor computational efficiency at inference (from repeated computations).

*Limitations of patch-based approach in fully convolutional mode:* A solution to improving computational efficiency during inference is the concept of Fully Convolutional Networks (FCN), which modifies a deep learning algorithm (more generally, an artificial neural network) to utilize only operations that are invariant to the input image size, such as convolutions and pooling. In this manner, a network designed to train with input of size 300 produces valid output even with larger input size, such as 5000 at inference. However, *valid* output does not imply *consistent* output (Extended Data Figure 2), defined as achieving the same overall output grid of predictions whether the network was used in FCN mode (5000 at inference) or not (300 at inference, applied with a sliding window to form the grid of outputs). In our example figure, naively applying the FCN results in grid-like artifacts that are not present when not leveraging the FCN. This artifact is caused by the popular 'same' padding option for convolutions, which preserves the input and output sizes to that operation by padding the input with an appropriate number of zeros at the border. These zeros do not cause issues when the training and inference patch sizes are the same, but application of the trained network in an FCN mode replaces the additional zeros with additional "context" from the image (Extended Data Figure 2). This mismatch in training and inference causes the grid-like artifacts.

*Proposed solution:* To solve this padding issue with FCNs applied to networks with 'same' padding convolutions, our proposed modification changes these paddings to 'valid' (no zero-padding) at training time, allowing an artifact-free FCN modification. As an additional contribution, we show that this is possible even with networks that branch into multiple pathways and merge using a channel-wise concatenation. Where each branch originally used valid padding to maintain the same spatial size for the concatenation, we modified this to crop the branches appropriately to the output size of the smallest branch (Extended Data Figure 2). This detail is important for the latest networks, many of which contain these branches[19].

*Other applications of the proposed approach:* Finally, we note that although we have presented this modified FCN approach as motivated by the image size of 5000x5000 pixels, the same concept applies (at a larger scale) for whole-slide scanner images, which are approximately 100,000x100,000 pixels. Usage of our approach can significantly speed up the application of deep learning algorithms to such large images.

**Deep Learning Algorithm Patch-based Training.** The deep learning image analysis workflow includes two phases: algorithm development and algorithm application, as illustrated in Extended Data Figure 2. For algorithm development, we trained the neural networks on digitized pathology slides with patches of size 911×911 pixels at each of magnifications (4X, 10X, 20X, 40X for lymph node and 4X, 10X, 20X for prostate), and the corresponding labels indicating the diagnosis, e.g. tumor/benign or Gleason grades. By changing the weights of the neural network to reduce the difference between the predicted results and the corresponding labels, the neural network learned to recognize patterns and distinguish between images with different labels. During neural network training, we also scaled the input (scanner) images by the appropriate factor to match the pixel resolution from the scanner images (~0.25μm/pixel) to the pixel resolution from the microscope camera (~0.11μm/pixel). The alternative of scaling the ARM image pixels in real time was rejected to minimize computation at inference in the ARM system. For algorithm application, the neural network was fed images of size 5120×5120 pixels captured from the microscope camera. The output from the network is a heatmap depicting the likelihood of cancer at each pixel location. The heatmap can be displayed directly using a colormap, or thresholded to get an outline that is then displayed as an overlay on the sample. ARM users favored these outlines over heatmaps to avoid occluding the view of the underlying sample. The ARM system is capable of displaying either visualization mode, and has the ability to quickly switch the augmented display off to examine the sample without deep learning assistance.

**Deep Learning Algorithm Evaluation.** We evaluate the algorithm performance for tumor detection within the field of view (FOV) with the following metrics: receiver operating characteristic (ROC) curves (the true positive rate against the false positive rate), area under the ROC curve (AUC), accuracy, precision, and recall (TP: true positive; TN: true negative; FP: false positive; FN: false negative):

$$Accuracy = (TP + TN) / (TP + TN + FP + FN)$$
$$Precision = TP/(TP+FP),$$
$$Recall\ or\ True\ Positive\ Rate = TP/(TP+FN),$$
$$False\ Positive\ Rate = FP/(FP+TN).$$

For each FOV, the output from the network is a heatmap depicting the likelihood of cancer at each pixel location. FOV likelihood was calculated by taking the maximum likelihood across all pixels. The FOV prediction was considered positive if the FOV likelihood was larger than equal to a chosen threshold, and FOV prediction was negative if otherwise. By sweeping the threshold from 0 to 1, we generated the ROC curve of the true positive rate against the false positive rate. The AUC is defined as the area under the ROC curve.

The FOVs for evaluation were collected to cover a diverse range of histological types. For the lymph node evaluation, the images contained lymphoid cells, connective tissue, blood or lymphatic vessels, fat, and metastatic breast cancer. For the prostate evaluation, the images contained benign prostatic glands, blood or lymphatic vessels, fat, inflammation, and prostate cancer of varying Gleason patterns. Note that the image quality from the microscope and whole slide scanner differ significantly with respect to the level of focus, exposure time, etc. In this study, we collect images with proper focus and exposure level for evaluation as in the regular microscopic tissue review workflow.

**Application agnostic platform.** The ARM is an application agnostic platform that operates based on the input pixel size of the pre-trained neural network. For networks with an input pixel size of at least 5120 pixels (configurable; this is the current ARM's image sensor resolution), the network will be used directly to make predictions. For pre-trained neural networks with an input pixel size smaller than 5120 pixels, the system will stride the network across the whole field-of-view, and assemble all the network inference results back into single field-of-view. A network that is fully convolutional can be run in either mode by either specifying the input size as 5120 (for a single-pass inference) or a smaller size for strided, "sliding-window" inference. Therefore, a single ARM add-on module works with any new convolutional neural network that follows this common format of image-in and prediction-out.

# Extended Data Figures

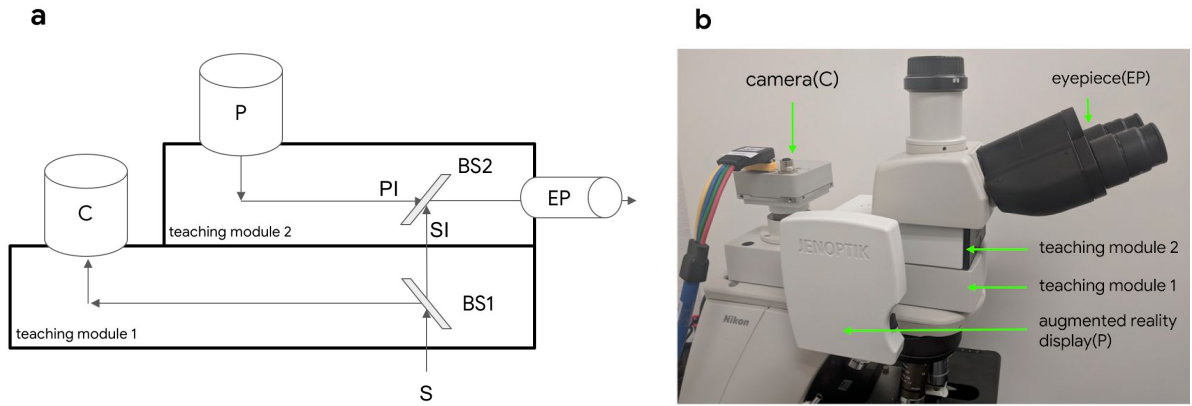

**Extended Data Figure 1 | System overview. a,** Schematic of the optic pathway. The standard upright microscope illuminates the specimen (S) from behind and captures the image rays with a conventional objective. These rays propagate upward, in a collimated state, towards the oculars. A teaching module (Nikon Y-IDP) with a beam splitter (BS1) was inserted into the optical pathway in the collimated light space. This module was modified to accept a microscope camera (C), so that the specimen image relayed from BS1 was in focus at the camera sensor when the specimen was also in focus to the microscope user. A second customized teaching module (Nikon T-THM) was inserted between the oculars and the first teaching module. The beam splitter in this module (BS2) was rotated 90 degrees to combine light from the specimen image (SI) with that from the projected image (PI) from the microdisplay (P). The augmented reality display includes a microdisplay and collimating optics, which were chosen to match the display size with the ocular size (22 mm). In this prototype, we tested two microdisplays: one that supports arbitrary colors (RGB), and another brighter display that supports only the green channel. The position of the collimator was adjusted to position the microdisplay in the virtual focal plane of the specimen. This collocation of SI and PI in the same plane minimizes relative motion when the observer moves, a phenomenon known as parallax. Note that BS1 needs to precede BS2 in the optical pathway from objective to ocular, so that camera C sees a view of the specimen without the projection PI. The observer looking through the eyepiece (EP) sees PI superimposed onto SI. **b,** Photograph of the actual implementation labeled with the corresponding modules.

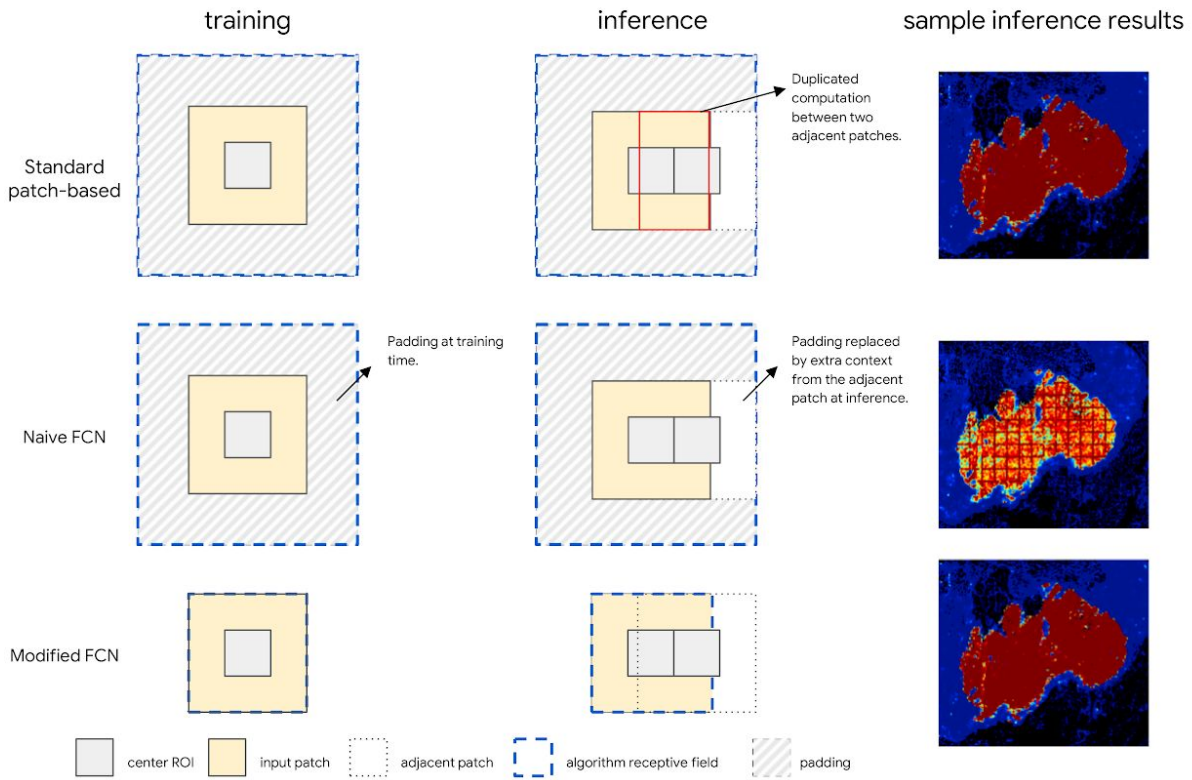

**Extended Data Figure 2 | Training and inference comparison across three design choices for the deep learning algorithm.** The standard patch-based approach crops the input image into smaller patches for training and applies the algorithm in a sliding window across these patches for inference. This results in poor computational efficiency at inference from repeated computations across adjacent patches. Naive FCN eliminates the adjacent computations but causes the grid-like artifacts due to the mismatched context between training and inference. Modified FCN removes paddings in the network ensuring consistent context between training and inference. This proves artifacts free inference results with no repeated computations between adjacent patches.

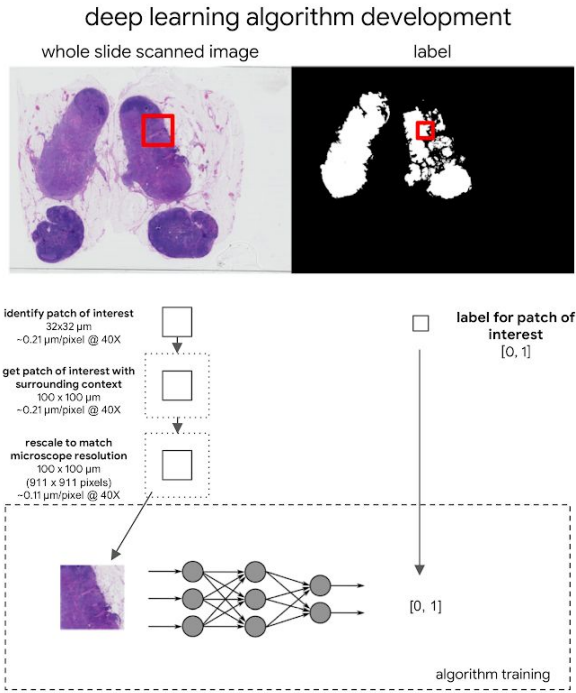
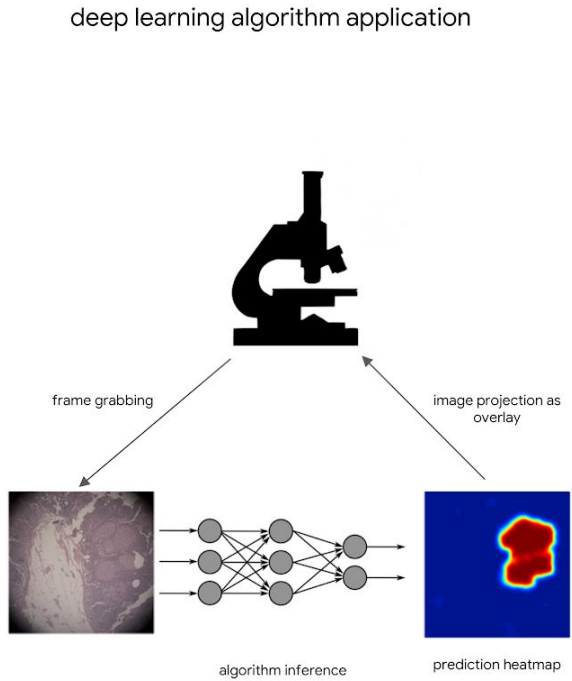

**Extended Data Figure 3 | Deep learning algorithm development and application.** In the development phase, we first sample patches with size (911×911 pixels) from digitized whole slide image. The patches are then preprocessed to match the data distribution of microscope images. In the application phase, an image with size (5120×5120 pixels) is provided to the network. The output of the network is a heatmap depicting the likelihood of cancer at each pixel location.

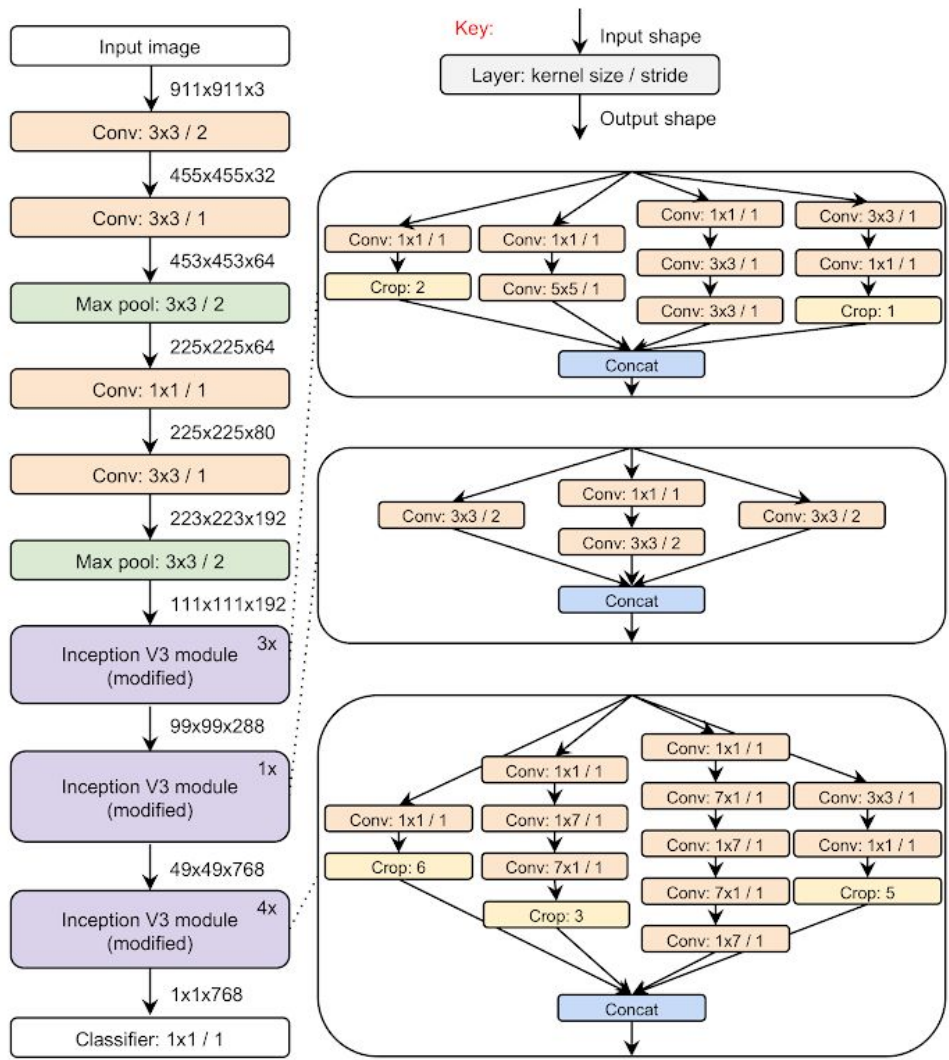

**Extended Data Figure 4 | Modified InceptionV3 network that avoids introducing artifacts when run in "fully convolutional mode" at inference.** "Crop" layers with parameter $k$ crop a border of width $k$ from its input. The principles we followed in the modifications were: **(1)** use of 'valid' instead of 'same' padding for all convolutions to avoid introducing artificial zeroes when the input size is increased at inference time; **(2)** differentially cropping the output of the branches in each Inception block as appropriate to maintain the same spatial size (height and width of each feature block) for the channel-wise concatenation operation; **(3)** increasing the receptive field to increase tissue context available for the neural network's interpretation.

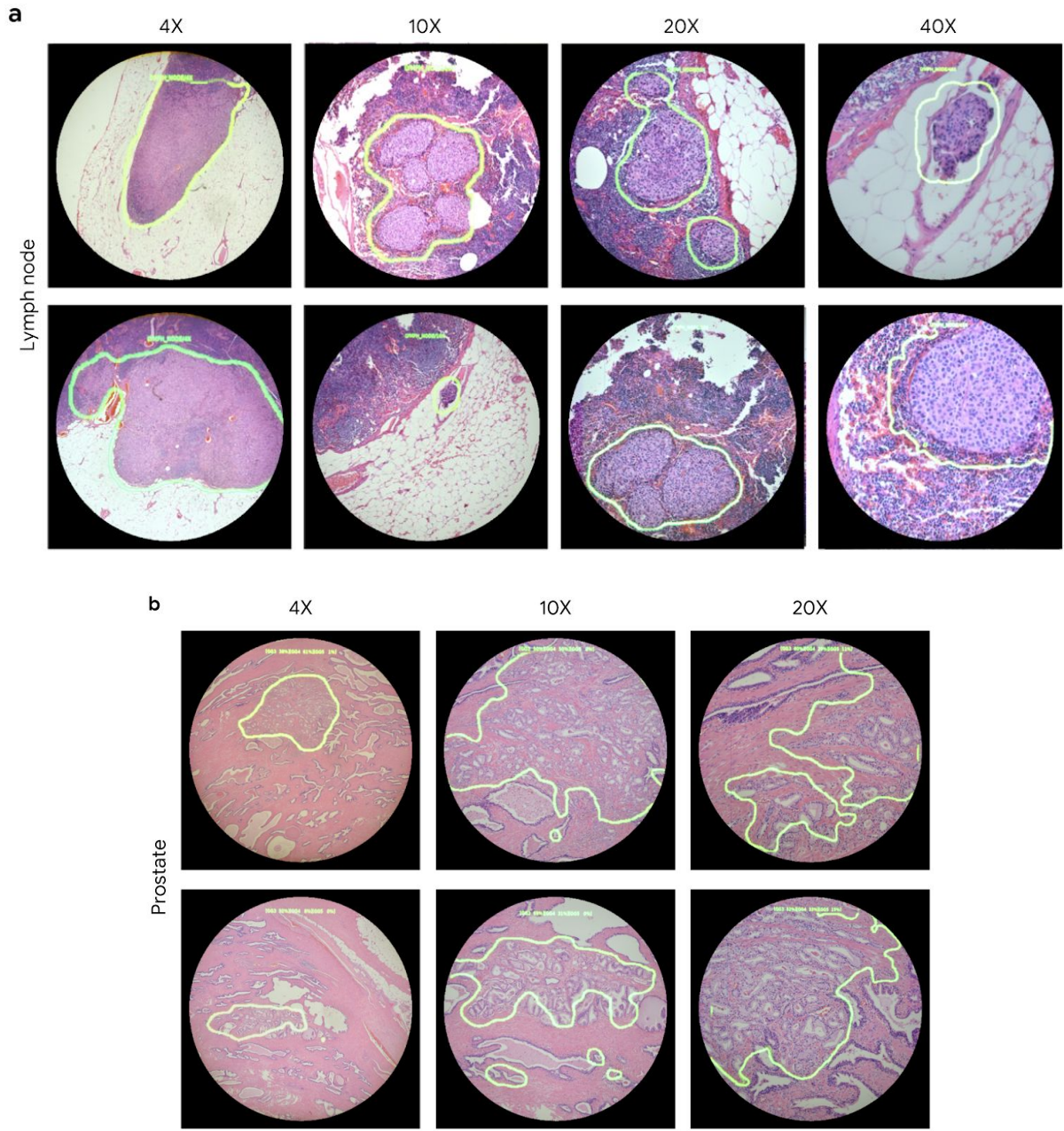

**Extended Data Figure 5 | Sample views through the lens.** The images show actual views through the lens of the ARM with green outlines highlighting the predicted tumor region. **a,** Left to right: lymph node metastases detection at 4X, 10X, 20X, and 40X. **b,** Left to right: prostate cancer detection at 4X, 10X, and 20X.

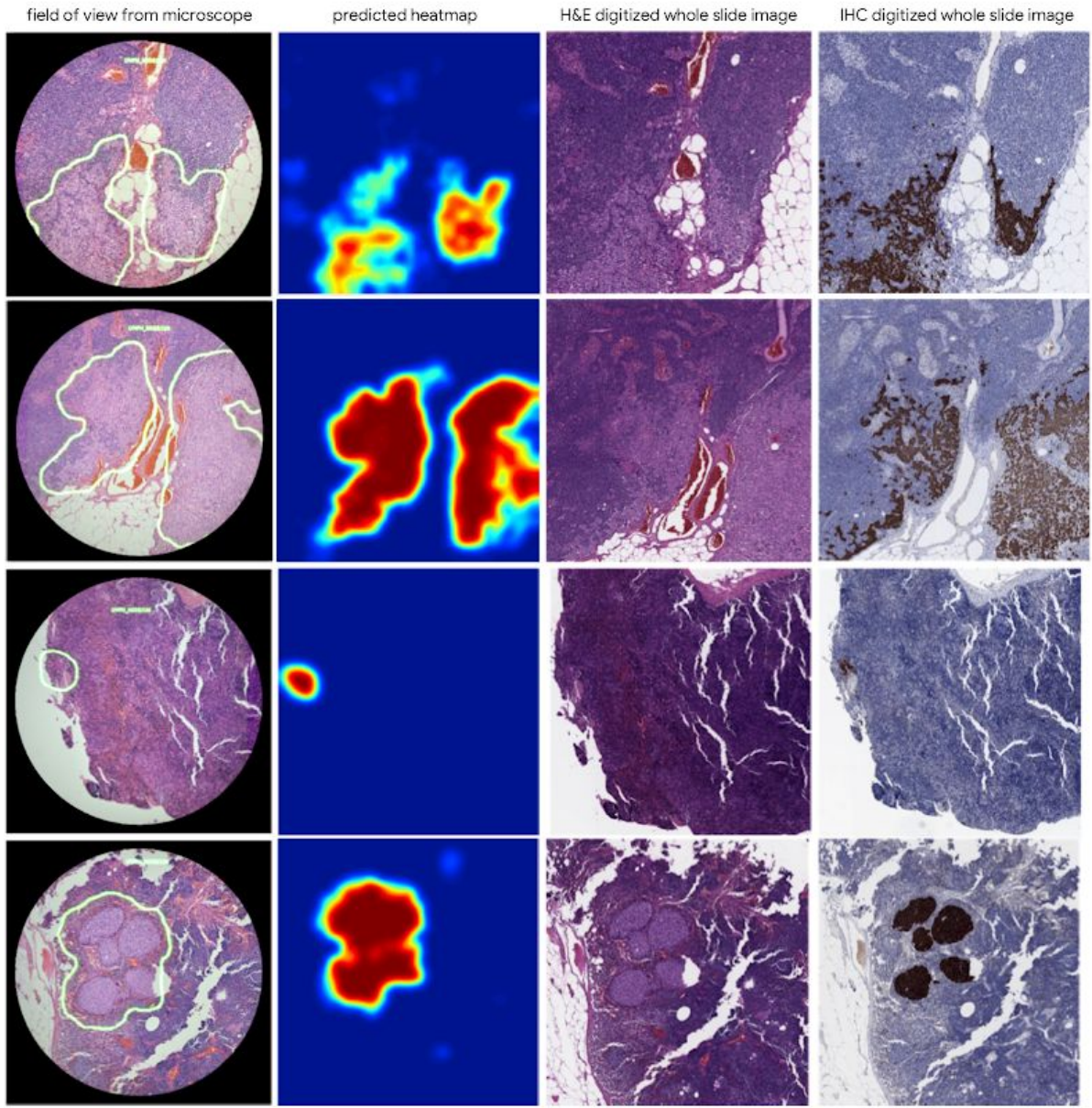

**Extended Data Figure 6 | Lymph node cancer detection at 10X compared with the corresponding immunohistochemistry (IHC) as the reference standard.** Left to right: Field of view as seen from the ARM, predicted heatmap, corresponding field of view from a digital scanner, corresponding field of view of the IHC (pancytokeratin AE1/AE3 antibody) stained slide from a digital scanner. This IHC stain highlights the tumor cells in brown.

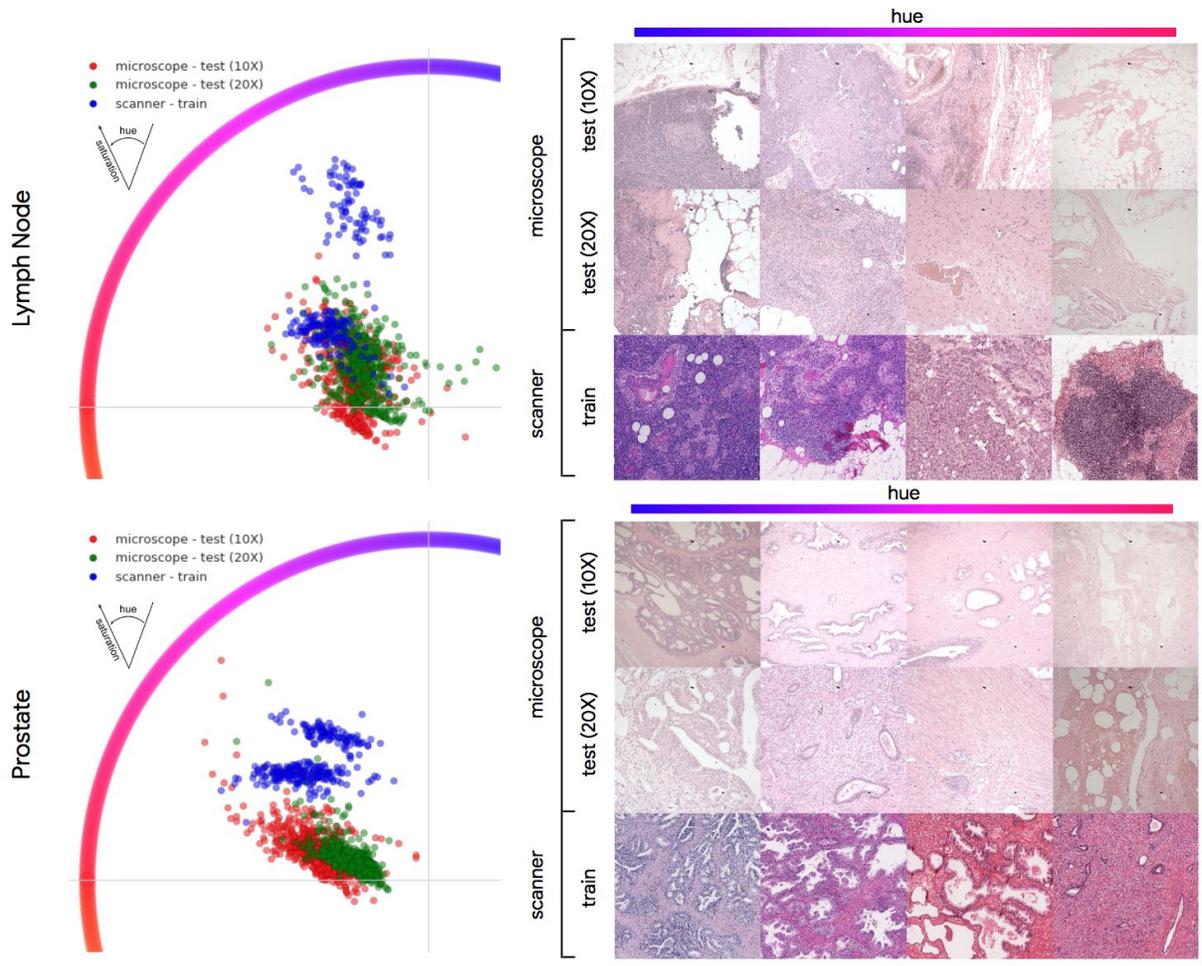

**Extended Data Figure 7 | Visualization of color distribution of slides used in the training and test sets.** In the polar scatter plots, the angle represents the hue (color), and the distance from origin represents the saturation. Each point represents the average hue and saturation of an image after mapping RGB values to optical densities followed by a Hue-Saturation-Density (HSD) color transform. The HSD transform is similar to Hue-Saturation-Value (HSV), but corrects for the logarithmic relationship between light intensity and stain amount, and has been shown to better represent stained slides[5,30]. The train set from digitized scanner is shown in blue and test sets from microscope are shown in red and green.

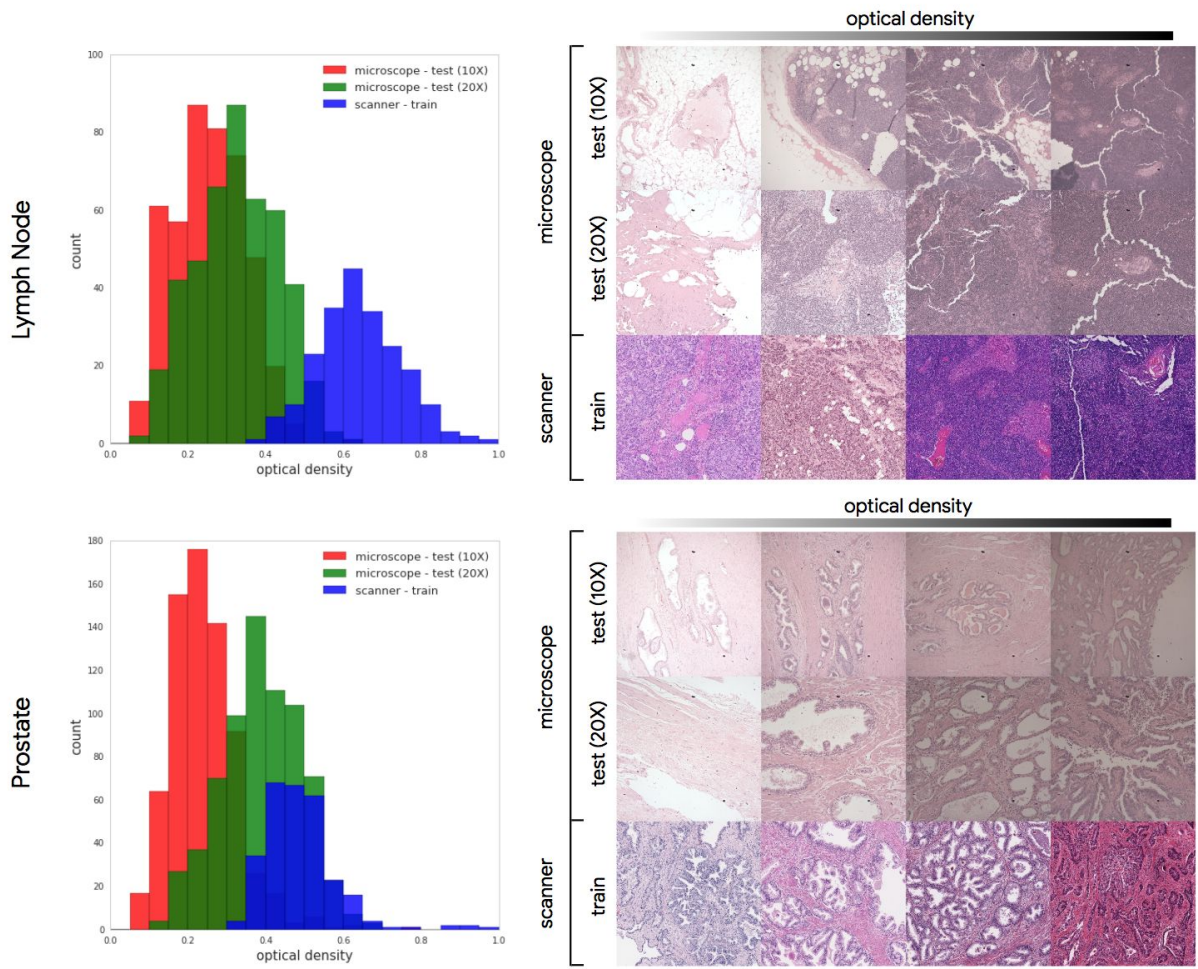

**Extended Data Figure 8 | Visualization of optical density distribution of slides used in the training and test sets.** In the histogram plots, the x axis represents the luma (brightness) of the image. Each point represents the average optical density of an image after mapping RGB values to optical densities followed by a Hue-Saturation-Density (HSD) color transform[30]. The train set from digitized scanner is shown in blue and test sets from microscope are shown in red and green.

| Lymph node test set | | | | | | |
|---|---|---|---|---|---|---|
| Category of metastatic tumor | Number of slides | | | Number of FOVs per magnification | | |
| | Data source 1 | Data source 2 | Total (%) | Benign | Tumor | Total (%) |
| Benign | 12 | 0 | **12 (24%)** | 120 | 0 | **120 (24%)** |
| Isolated Tumor Cells | 3 | 5 | **8 (16%)** | 70 | 10 | **80 (16%)** |
| Micrometastasis | 4 | 14 | **18 (36%)** | 138 | 42 | **180 (36%)** |
| Macrometastasis | 8 | 4 | **12 (24%)** | 76 | 44 | **120 (24%)** |
| **Total** | **27** | **23** | **50 (100%)** | **404** | **96** | **500 (100%)** |

| Prostate test set | | | | | | |
|---|---|---|---|---|---|---|
| Gleason grade group (GG) | Number of slides | | | Number of FOVs per magnification | | |
| | Data source 1 | Data source 2 | Total (%) | Benign | Tumor | Total (%) |
| Benign | 6 | 0 | **6 (18%)** | 120 | 0 | **120 (18%)** |
| GG1 | 10 | 0 | **10 (29%)** | 143 | 57 | **200 (29%)** |
| GG2-3 | 7 | 0 | **7 (21%)** | 80 | 60 | **140 (21%)** |
| GG4-5 | 1 | 10 | **11 (32%)** | 114 | 106 | **220 (32%)** |
| **Total** | **24** | **10** | **34 (100%)** | **457** | **223** | **680 (100%)** |

**Extended Data Table 1 | Number and breakdown of slides and fields of view in the test datasets.** The test sets contain slides from two data sources to improve the diversity of categories represented. Each slide was drawn from a distinct patient's case. The reference standard labels for the lymph node test set was established by reviews from two pathologists and adjudication by a third, using cytokeratin immunohistochemistry staining for reference. The reference standard labels for the prostate test set was established by reviews from three pathologists, using PIN4 immunohistochemistry staining where available.

| Tissue Type | Magnification | AUC (CI$_{95}$) | Operating Threshold | Accuracy (CI$_{95}$) | Precision (CI$_{95}$) | Recall (CI$_{95}$) |
|---|---|---|---|---|---|---|
| lymph node | 10X | **0.96 (0.93-0.98)** | High Accuracy | **0.95 (0.93-0.97)** | 0.99 (0.95-1.00) | 0.74 (0.66-0.83) |
| | | | High Precision | 0.94 (0.92-0.96) | **0.99 (0.95-1.00)** | 0.69 (0.60-0.78) |
| | | | High Recall | 0.77 (0.73-0.81) | 0.46 (0.39-0.53) | **0.95 (0.90-0.99)** |
| | 20X | **0.98 (0.96-0.99)** | High Accuracy | **0.95 (0.93-0.97)** | 0.94 (0.88-0.99) | 0.78 (0.70-0.86) |
| | | | High Precision | 0.93 (0.91-0.95) | **0.98 (0.95-1.00)** | 0.67 (0.57-0.76) |
| | | | High Recall | 0.85 (0.82-0.88) | 0.57 (0.50-0.65) | **0.96 (0.91-0.99)** |
| prostate | 10X | **0.95 (0.94-0.96)** | High Accuracy | **0.99 (0.89-0.93)** | 0.79 (0.74-0.84) | 0.99 (0.97-1.00) |
| | | | High Precision | 0.93 (0.91-0.95) | **0.83 (0.79-0.88)** | 0.97 (0.95-0.99) |
| | | | High Recall | 0.88 (0.86-0.91) | 0.74 (0.69-0.79) | **1.00 (0.99-1.00)** |
| | 20X | **0.98 (0.97-0.99)** | High Accuracy | **0.96 (0.94-0.97)** | 0.94 (0.91-0.97) | 0.92 (0.89-0.95) |
| | | | High Precision | 0.96 (0.94-0.97) | **0.94 (0.91-0.97)** | 0.92 (0.89-0.95) |
| | | | High Recall | 0.92 (0.90-0.94) | 0.80 (0.75-0.85) | **0.99 (0.98-1.00)** |

**Extended Data Table 2: Performance metrics for lymph node metastases detection and prostate cancer detection.** The table shows AUC and accuracy, precision and recall with three different operating thresholds: a high accuracy operating threshold, a high precision operating threshold for a diagnostic use case, and a high recall threshold for a screening use case. The operating point can be configured during usage. Confidence intervals (CIs) were calculated with 5000 bootstrap replications.

# Supplementary Information

## Supplementary Video

https://youtu.be/9Mz84cwVmS0

**Extended Video 1: Overview of the Augmented Reality Microscope.**